# MOB-ESP and other Improvements in Probability Estimation

**Rodney D. Nielsen**
Dept of Computer Science
University of Colorado
Boulder, CO 80309-0430
Rodney.Nielsen@Colorado.edu

## Abstract

A key prerequisite to optimal reasoning under uncertainty in intelligent systems is to start with good class probability estimates. This paper improves on the current best probability estimation trees (Bagged-PETs) and also presents a new ensemble-based algorithm (MOB-ESP). Comparisons are made using several benchmark datasets and multiple metrics. These experiments show that MOB-ESP outputs significantly more accurate class probabilities than either the baseline B-PETs algorithm or the enhanced version presented here (EB-PETs). These results are based on metrics closely associated with the average accuracy of the predictions. MOB-ESP also provides much better probability rankings than B-PETs. The paper further suggests how these estimation techniques can be applied in concert with a broader category of classifiers.

## 1 INTRODUCTION

The standard form of supervised machine learning involves inducing a classifier $f$ from a set of labeled training examples $\langle \mathbf{x}_i, y_i \rangle$, where each $\mathbf{x}_i \in X$ is a vector composed of several predictive attributes and $y_i \in \{1, 2, ..., K\}$ is the class label to be predicted. The goal is to induce a classifier which minimizes the expected error rate resulting from the zero-one loss between the predicted label $\hat{y} = f(\mathbf{x})$ and the true label $y$.

In many circumstances, it is not enough to simply have a predicted class label, $\hat{y}$. We would also like a prediction of the conditional probability of each class $k$, $\hat{p}(y{=}k|\mathbf{x})$, henceforth written simply as $\hat{p}(k|\mathbf{x})$. Consider the medical domain, where the certainty of a diagnosis directly effects whether additional tests, possibly invasive tests, should be carried out or what treatment protocol should be followed. In making other cost-sensitive decisions, it is also important to understand the likelihood of each possible outcome (Margineantu and Dietterich, 2000; Piatetsky-Shapiro and Masand, 1999). That is, the cost or benefit associated with acting on various class predictions might be significantly different and high-quality decision making requires good class probability estimates.

The focus of this paper is on this probabilistic classification task. More specifically, it analyzes probability estimation techniques based on ensemble decision tree classifiers. Frequently, this is done by estimating class probabilities $\hat{p}(k|\mathbf{x})$ at each leaf in a tree based on the proportion of training examples from class $k$ that reach the leaf (i.e., $\hat{p}(k|\mathbf{x}) = n_k/n$, where $n$ is the total number of training examples that reach the leaf and $n_k$ is the number of those examples from class $k$). These predictors are called Probability Estimation Trees (PETs).

Margineantu and Dietterich (2002) point out three problems with this approach. First, since the estimates are based on the same training data used to build the tree, they tend to be too close to the extremes of 1.0 and 0.0. Second, additional inaccuracies result from the small number of examples that often reach a leaf. Third, this technique assigns the same probability to the entire region of space defined by a given leaf without regard for the true probability density function across that region.

Generally these estimates are smoothed to mitigate the problems caused due to few examples reaching many of the leaves (Bauer and Kohavi, 1999; Provost and Domingos, 2003). The simplest form of smoothing is Laplace smoothing, where the probability estimate at a given leaf is calculated as

$$\hat{p}(k|\mathbf{x}) = \frac{n_k + 1}{n + K}, \qquad (1)$$

where $K$ is the number of classes. An alternative form of smoothing is the $M$-estimate (Cestnik, 1990)

$$\hat{p}(k|\mathbf{x}) = \frac{n_k + p_k m}{n + m}, \qquad (2)$$

where $p_k$ is the prior or base rate for class $k$ and $m$ is a smoothing constant. Often $p_k$ is set to $\frac{1}{K}$, leading to Laplace smoothing when $m = K$.

Probabilities can also be calculated using ensemble forecasting techniques (c.f. Toth, 2002). Here the class $k$ probability estimate is determined by the percentage of deterministic classifiers in the ensemble that vote for class $k$, (i.e., $\hat{p}(k|\mathbf{x}) = T_k/T$, where $T$ is the total number of classifiers in the ensemble and $T_k$ is the number of those classifiers that output $\hat{y}=k$). To work reasonably well, this technique requires that the underlying classifiers be independent and that the ratio of classifiers to classes $T/K$ be relatively large. For example, suppose $K$=25, even if $T$=100, it is likely that several classes will be assigned a zero probability. As with PETs, smoothing could be used to mitigate this problem.

Bagging (Breiman, 1996) further improves the smoothed estimates of PETs. In a sense, this technique combines the ideas of PETs and ensemble forecasting. In Bagged-PETs (B-PETs[1]), the estimates output by each classifier in the ensemble are averaged to provide a final prediction (Bauer and Kohavi, 1999; Margineantu and Dietterich, 2002; Provost and Domingos, 2003). Bagging can be seen as partially addressing the problem of region-wide probability estimates in that two unlabeled examples, $\mathbf{x}_u$ and $\mathbf{x}_{u'}$, are unlikely to be classified via the same leaf in every tree and will, therefore, generally have different final probability estimates.

Margineantu and Dietterich (2002) further lessen the region-wide estimation problem and improve the probability estimates using B-LOTs, a modification that includes lazy decision tree learning (Friedman et al., 1996) and Options (Buntine, 1990; Kohavi and Kunz, 1996). Lazy decision trees are built as a single branch at the time of classifying unlabeled example $\mathbf{x}_u$. This branch is constructed based partly on $\mathbf{x}_u$ and is relatively specific to $\mathbf{x}_u$. Therefore, the probability estimate is more specific to the point $\mathbf{x}_u$, rather than a broader region that just includes $\mathbf{x}_u$ somewhere within its boundaries. Option trees involve multiple split decisions at each node with final probabilities computed by averaging over the results from all paths. This increases the diversity of the ensemble and improves the subsequent probability estimates. As pointed out by Margineantu and Dietterich, B-LOTs require more computational resources during classification and are, therefore, inappropriate in many situations. Furthermore, while B-LOTs provide more accurate probability estimates (estimates with lower mean squared error), they do not provide nearly as good probability rankings as do B-PETs.[2]

Ferri et al. (2003) present a new smoothing algorithm *m-Branch smoothing* that results in some improvement of probability rankings in single-tree PETs. However, it was not directly evaluated on error in predicted probabilities.

The goal in this paper is to develop an algorithm that performs extremely well on the average prediction accuracy. Therefore, the benefits of single trees (e.g., interpretability) are disregarded in favor of the potential accuracy improvements provided by ensembles. Additional goals include high accuracy of probability rankings, relatively low computational costs at classification time, and reasonable performance on small datasets. For these reasons, B-PETs, rather than B-LOTs, is implemented for comparison purposes in this paper.

Two new algorithms are presented, an enhanced version of B-PETs (EB-PETs), and an algorithm that uses an entirely different strategy to predict conditional class probabilities (MOB-ESP). The work here builds most directly on the work of Breiman (2001), Margineantu and Dietterich (2002), and Provost and Domingos (2003), though the work of many others has been influential. These algorithms are evaluated using 20 benchmark datasets and multiple evaluation metrics used extensively in the literature. The experiments show that both EB-PETs and MOB-ESP significantly outperform the baseline B-PETs algorithm on all metrics. They also show that MOB-ESP performs much better than EB-PETs in regard to the average accuracy of its class probability predictions.

In section 2, we describe the algorithms evaluated in this paper, including B-PETs, EB-PETs, and MOB-ESP. Section 3 presents the experiments, including subsections on their design, datasets, evaluation metrics, results, and discussion. Concluding remarks are provided in section 4.

## 2 ALGORITHMS

### 2.1 PROBABILITY ESTIMATION TREES

As the baseline for comparison, Bagged PETs (B-PETs) was implemented as described in (Provost and Domingos, 2003). They use the basic C4.5 algorithm (Quinlan, 1993), but with pruning and collapsing disabled and with Laplace smoothing as describe in equation 1 to estimate class probabilities at the leaves. Turning off all pruning, the tree is grown to purity, but must contain at least 2 examples in each leaf. Bagging is added to complete B-PETs and the final probability estimate for a class $k$ is calculated by taking the average of the $T$ estimates output by the trees of the ensemble $\hat{p}_t(k|\mathbf{x})$,

$$\hat{p}(k|\mathbf{x}) = \frac{1}{T}\sum_{t=1}^{T}\hat{p}_t(k|\mathbf{x}). \quad (3)$$

Margineantu and Dietterich (2002) point out that, while smoothing addresses the issue of small leaves, and Bagging combined with no pruning addresses the issue of region-wide probability estimation, B-PETs does not address the problems caused by basing estimates on the training data – chiefly, that estimates tend to be biased toward the extremes of 0.0 and 1.0. They attempted to address that problem by calculating $\hat{p}_t(k|\mathbf{x})$ based only on *out-of-bag* (OB) examples (those training points not used in the construction of the associated tree), but found that there were too few OB examples to generate meaningful estimates.[3] This training data issue is more directly addressed by MOB-ESP, described in section 2.3.

---

[1]Bagged PETs are referred to in the literature as p-Bagging (Bauer and Kohavi, 1999), C4.4-B (Provost and Domingos, 2003) and B-PETs (Margineantu and Dietterich, 2002).

[2]A predictor with perfect probability rankings is characterized by having estimates that increase monotonically with increases in the true probabilities (c.f., Margineantu and Dietterich, 2002).

The algorithm here differs from the implementation of Provost and Domingos in two minor ways. First, rather than follow the standard bootstrap selection procedure, *in-bag* (IB) examples (those training points randomly chosen to construct a given tree) are drawn according to their training set priors. That is, normally the examples used to build a bagged classifier are a bootstrap replicate of the full training dataset; $N$ training examples are drawn uniformly and with replacement, where $N$ is the size of the full training set (Efron and Tibshirani, 1993). Here, for each class $k$, $N_k$ class $k$ examples are chosen in this same way, where $N_k$ is the number of class $k$ examples in the full training set. Second, if two or more attributes result in the same information gain, the branching decision (split) is determined by uniform random selection among those attributes.

## 2.2 ENHANCED B-PETS

B-PETs is enhanced in three ways to form EB-PETs. Each enhancement is described and implemented separately to allow individual analysis. First, OB examples are included in the estimation of probabilities. Second, the Laplace smoothing is eliminated from B-PETs. Third, random feature selection is incorporated into the tree construction process. These enhancements are detailed successively in the following three sections.

### 2.2.1 Including Out-of-Bag Examples

The first improvement made to B-PETs is to include the OB examples in the probability estimation process. These are the examples that are stochastically left out during the $t$th bootstrap replicate generation and are, therefore, not used to construct tree $t$. After constructing the tree, these OB examples are classified via the tree and binned in the leaves they reach. The Laplace smoothed probability estimates for tree $t$ are now calculated as follows:

$$\hat{p}_t(k|\mathbf{x}) = \frac{n_k^{IB} + \alpha n_k^{OB} + 1}{n^{IB} + \alpha n^{OB} + K}, \quad (4)$$

where the superscripts $IB$ and $OB$ represent those training examples used to create tree $t$ and those that are not,

respectively, and $\alpha$ is a constant used to weight the OB examples more heavily (in the implementation here, $\alpha$=1). Otherwise, the notation is the same as that of equation 1.

The hypothesis is that including the OB examples will further mitigate two of the problems with PETs described previously. First, this should alleviate the problems of basing probabilities strictly on the examples used to construct the tree. Now, approximately 37% of the examples will not have been involved in constructing the tree (see footnote 3[4]). Second, including the OB examples increases the size of leaves. Effectively, $n$ and $n_k$ from equation 1 are increased by about 58%.[5] To the best of my knowledge, this technique has not been implemented in prior work.

### 2.2.2 Omission of Smoothing

Various prior work has shown that unsmoothed estimates can circumstantially achieve better accuracy. For example, Margineantu and Dietterich (2002) did not use smoothing in B-LOTs, because the Options aspect of the algorithm implicitly provided some smoothing. Wondering whether the same is true for Bagging alone, a version of the algorithm was implemented without smoothing

$$\hat{p}_t(k|\mathbf{x}) = \frac{n_k^{IB} + \alpha n_k^{OB}}{n^{IB} + \alpha n^{OB}}. \quad (5)$$

Additionally, it was hypothesized that the 58% increase in probability estimation points due to including the OB examples and the diversity injected by random feature selection (described next) would further decrease the need for explicit smoothing.

### 2.2.3 Random Feature Selection

The last algorithm enhancement incorporates random feature selection (Amit and Geman, 1997; Breiman, 2001) into the tree building process. At each node in the tree, only a random subset of attributes are considered in making the split. Specifically, we examine $d$ attributes according to the following formula

$$d = \lceil \sqrt{D} \rceil, \quad (6)$$

where $D$ is the total number of attributes in the dataset.

In some sense, random feature selection is addressing the same issues as Options (Buntine, 1990; Kohavi and Kunz, 1996; Margineantu and Dietterich, 2002). Margineantu and Dietterich indicate the advantage of Options is that splits with an information gain almost as high as the best test will be created, where this is not guaranteed in decision trees or bagged trees. This leads to more diversity and better probability estimates. Random feature selection has this same positive effect, compounded by the relatively small number of attributes evaluated at each node in the tree.

---

[3]The probability of a training example not being selected in the bootstrap replicate process and, therefore, being an OB example is $\left(\frac{N-1}{N}\right)^N \approx \frac{1}{e} \approx 0.37$, for even modest $N$.

[4]Technically, in this case, we should use $N_k$, where the examples are from class $k$, but this has little effect for modest $N_k$.

[5]The number of examples used increases from $n^{IB} \approx (1 - 0.37)N$ to $N$, an increase of about $\frac{N-(1-0.37)N}{(1-0.37)N} \approx 0.58$.

The hypothesis is that this technique will lessen two of the problems described earlier. First, it should provide additional implicit smoothing by injecting randomness into the makeup of leaves. This will mitigate the problems caused by small leaves. Second, this diversity will also further address the issue of region-wide probability estimates, since it becomes even less likely that the regions of input space covered by leaves of different members of the ensemble will overlap greatly. More specifically, it should be less probable that any two unlabeled examples will reach and be classified by the exact same set of leaves across the entire ensemble.

### 2.3 MOB-ESP

Several of the previous modifications attempt to alleviate the problems caused by using the same data points to both build the tree and estimate the probabilities. The algorithm described here, Mean Out-of-Bag Example-Specific Probability estimator (MOB-ESP), more explicitly addresses this issue. It does so by using the true and predicted class of OB examples to estimate conditional class probabilities of unlabeled examples that fall in their region of the input space.

Constructing a MOB-ESP is a three step process. First, a bagged tree classifier with random feature selection is constructed as described in previous sections. During each tree's construction, the set of in-bag examples that compose a leaf $\Theta_l^{IB}$ are binned with that leaf. Additionally, the OB examples are passed through the tree during (or after) construction and the set of OB examples classified via a leaf $\Theta_l^{OB}$ are binned separately with the leaf.

Second, after all of the trees in the ensemble are constructed, the entire training dataset is classified via their OB trees using the standard majority voting scheme from Bagging. This results in the set of classifications $\{\langle \mathbf{x}_i, \hat{y}_i \rangle\}_{i=1}^N$. For a training point $\mathbf{x}_i$, the *OB trees* are the trees in which $\mathbf{x}_i$ is an OB example. On average, a training example $\mathbf{x}_i$ is an OB example in about $0.37T$ trees, where $T$ is the number of trees in the ensemble (see footnote 3). Since these classifications are based on OB trees, they are unbiased in the sense that the examples were not used in the construction of the trees employed to classify them and, therefore, their classification performance should be almost indistinguishable from that of unlabeled test examples. These classifications will be used subsequently to estimate the class probabilities for similarly classified test examples.

The final step in the MOB-ESP construction is to assign probabilities $\hat{p}_l(k|\hat{y})$ to each leaf of every tree in the ensemble. These probabilities are conditioned on the ensemble classification $\hat{y}$ of a point under the rationale that points classified similarly will have more similar true class probabilities than examples with different classifications. The motivation for this hypothesis is that, given a group of examples falling in the same leaf of a tree, those examples with the same ensemble classification are more likely to be in contextually similar regions of space (i.e., they are more likely to have been classified by a larger number of the same leaves across the ensemble). The hypothesis was tested empirically and found to be true for all of our benchmark datasets (numbers not reported, due to space). Now, the estimated class probabilities, given an ensemble classification $\hat{y}=j$, are computed according to the true class frequencies of the training examples binned with the leaf and having the classification $j$. In other words, to compute the class probability estimates for a point with classification $j$, the set of binned IB and OB training examples, $\Theta_l^{IB}$ and $\Theta_l^{OB}$, are processed as follows. Filter the sets to include only the points whose classification was $j$, i.e., compute

$$\Theta_{l,j}^{IB} = \{\mathbf{x}_1, \mathbf{x}_2, ...\mathbf{x}_{M^{IB}}\}, \qquad (7)$$

where $\mathbf{x}_m \in \Theta_l^{IB}$ and $f(\mathbf{x}_m)=j$. The set $\Theta_{l,j}^{OB}$ is computed similarly using the OB training examples binned at the leaf. The conditional class $k$ probability predicted by leaf $l$ is then calculated as follows

$$\hat{p}_l(k|\hat{y}=j) = \frac{M_k^{IB} + \alpha_l M_k^{OB}}{M^{IB} + \alpha_l M^{OB}}, \qquad (8)$$

where $M^{IB}$ is the total number of IB training examples in $\Theta_{l,j}^{IB}$; $M_k^{IB}$ is the number of those same examples whose true class is $k$; $M^{OB}$ and $M_k^{OB}$ have parallel definitions based on the OB training examples in $\Theta_{l,j}^{OB}$; and $\alpha_l$ is a weighting factor to increase the role of the OB examples (in this implementation, $\alpha_l=1$). The result of this process is a $K \times K$ matrix of predicted class probabilities conditioned on the possible classifications.

To estimate the class probabilities of an unlabeled example $\mathbf{x}_u$, the point is first classified by the ensemble using the majority voting scheme of Bagging to obtain $f(\mathbf{x}_u)=\hat{y}_u$. Then for each leaf involved in the classification of $\mathbf{x}_u$, the conditional class $k$ probability $\hat{p}_l(k|\hat{y}_u=j)$ is retrieved from the leaf's probability matrix. Finally, the ensemble probability prediction for class $k$ is computed by averaging the estimates from all of the trees

$$\hat{p}(k|\mathbf{x}_u) = \frac{1}{T}\sum_{t=1}^{T} \hat{p}_t(k|\mathbf{x}_u), \qquad (9)$$

where $\hat{p}_t(k|\mathbf{x}_u) = \hat{p}_l(k|\hat{y}_u=j)$,[6] the conditional class $k$ probability calculated by the classifying leaf $l$ in tree $t$. In practice, the sets $\Theta_{l,j}^{IB}$ and $\Theta_{l,j}^{OB}$ might be empty for some leaves $l$. These leaves are not included in the averaging.

MOB-ESP is now cast in the light of the three problems outlined for PETs by Margineantu and Dietterich

---

[6]This implementation does not handle unknown attribute values; if it did, $\hat{p}_t(k|\mathbf{x})$ would be computed as a weighted sum of the results of multiple leaves instead of the strict equivalence here.

(2002). First, MOB-ESP significantly mitigates problems that might be caused by basing probability estimates on the data used to construct the trees. This is as a result of two factors: 37% of the examples used to calculate the estimates (the OB examples) are not used in the construction of the tree; and, even for the examples that are used in the construction, the estimates are not determined strictly by the training class labels, but also by their OB tree ensemble classifications – these are effectively non-training classifications. Second, the problem of small leaf sizes is mitigated, as in EB-PETs, by two factors: the 58% increase from including OB examples and the implicit smoothing resulting from Bagging and random feature selection. Third, the problem of region-wide estimation is alleviated by Bagging, random feature selection, and by conditioning the estimates on the ensemble classifications.

## 3 EXPERIMENTS

### 3.1 DESIGN AND DATASETS

The experiments compare each algorithm using 20 benchmark UCI datasets, as indicated in table 1.[7] One hundred trials were conducted for each dataset holding out a random 1/3 of the examples for test. Each trial ensemble consisted of 128 trees. This resulted in a total of 2000 ensembles and 256,000 total trees being utilized for each algorithm.

While the formulas throughout this paper are written for the general $K$-class scenario, the current implementation is restricted to the 2-class case. For multi-class datasets, the predictors were constructed using only two of the classes, which are indicated in the dataset name column.[8]

### 3.2 EVALUATION METRICS

We analyze the results based on multiple metrics for a variety of reasons. There is no consensus among researchers that any one metric is best; different metrics each have their own strengths and weaknesses (Toth et al., 2002; Zadrozny and Elkan, 2001). Presenting multiple metrics allows others to more easily compare to these results. The first two metrics are closely linked with the average error on predictions and the third and, to some extent, fourth are associated

Table 1: Datasets: classes used appear in parentheses, N = number of examples, D = number of attributes

| Id | Dataset | N | D |
|----|---------|-----|-----|
| 1 | audiology (3,8) | 92 | 37 |
| 2 | bupa | 345 | 6 |
| 3 | credit-german | 653 | 44 |
| 4 | glass (1,4) | 79 | 9 |
| 5 | iondata | 351 | 34 |
| 6 | iris (1,2) | 100 | 4 |
| 7 | letter (8,11) | 1473 | 16 |
| 8 | lymphography (2,3) | 142 | 18 |
| 9 | pima | 768 | 8 |
| 10 | segment (3,5) | 660 | 19 |
| 11 | sonar | 208 | 60 |
| 12 | soybean (14,15) | 80 | 35 |
| 13 | vehicle (1,2) | 429 | 18 |
| 14 | votes | 232 | 16 |
| 15 | vowel (5,6) | 180 | 12 |
| 16 | wbc | 683 | 9 |
| 17 | wdbc | 569 | 30 |
| 18 | wine (1,2) | 130 | 13 |
| 19 | wpbc | 194 | 33 |
| 20 | zoo (1,2) | 61 | 16 |

more closely with the quality of probability rankings.

The first metric is the mean squared error on the predicted probability of the correct class $y_u$ (Bauer and Kohavi, 1999; Provost and Domingos, 2003)

$$0/1\text{-MSE} = \tfrac{1}{U} \sum_{u=1}^{U} (1 - \hat{p}(y_u|\mathbf{x}_u))^2, \quad (10)$$

where $U$ is the size of the test set. In this 2-class case, 0/1-MSE is exactly one half the MSE metric or Brier score.

The second metric used is the average log-loss (AvLL) (Zadrozny and Elkan, 2001).[9] For the two-class case where the true probabilities are unknown, this is defined as

$$\text{AvLL} = -\tfrac{1}{U} \sum_{u=1}^{U} \log_2 \hat{p}(y_u|\mathbf{x}_u). \quad (11)$$

The third metric is the area under the lift chart (AULC) (Piatetsky-Shapiro and Masand, 1999; Zadrozny and Elkan, 2001).[10] Lift $l$ is computed by first sorting the predictions for class $k$ in descending order. Then for a proportion $v$ of the examples with the highest probability predictions $\hat{p}(k|\mathbf{x}_u)$, the lift is

$$l(v) = \tfrac{r_v}{r_{1.0}}, \quad (12)$$

where $r_v$ is the fraction of those examples that are actually from class $k$, and $r_{1.0}$ is the fraction of all test points that are from class $k$. AULC is calculated by numeric integration of $l(v)|_0^1$. Often this is done in increments of 0.05, but the implementation here was point-by-point across the uniquely ranked probability estimates. Typically, class $k$ in this scenario is the group that has a higher value associated with their identification (e.g., people who will buy a product). In most of the datasets here, there is no clear class $k$, so an $\text{AULC}_k$ was computed for each class and AULC was calculated by weighting these values by their associated prior class probabilities according to the training set.

---

[7] The implementation here did not explicitly accommodate unknown feature values or nominal values. Therefore, datasets were chosen to minimize any associated problems. Dataset examples with missing attribute values were removed. Audiology attributes *bone* and *bser* were removed, due to their large number of missing values. Nominal attribute values were converted to integers in accordance with a fairly natural progression (e.g., the Audiology *speech* attribute values normal, good, very_good, very_poor, poor, and unmeasured were converted to 0, 1, 2, -3, -2, and -1, respectively). Exact datasets available upon request.

[8] The two classes were selected based on a combination of having a higher number of examples and/or having a higher confusion rate than other classes.

[9] Unsmoothed estimates occasionally lead to 0/1 probability predictions, which are transformed to $\epsilon$/1-$\epsilon$ to avoid infinite log-loss values (in the experiments here, $\epsilon = \min(0.005, \min(0.5\hat{p}(k)))$ ).

[10] AULC is used rather than the area under the ROC curve (AUC), because AULC is employed more widely in commercial probability estimation and because AUC fails to distinguish between properly ranked estimates and estimates that are both properly ranked and have a low error (Zadrozny and Elkan, 2001).

The final metric is the change in classification accuracy ($\Delta\text{Acc}$). Each test point $\mathbf{x}_u$ is reclassified based on its highest probability class

$$f'(\mathbf{x}_u) = \hat{y}'_u = \text{argmax}_k \ \hat{p}(k|\mathbf{x}_u). \quad (13)$$

$\Delta\text{Acc}$ is then calculated as the change in the mean test set accuracy using the standard zero-one loss formulation.

$$\Delta\text{Acc} = \tfrac{1}{U}\sum_{u=1}^{U}\delta(y_u, \hat{y}'_u) - \tfrac{1}{U}\sum_{u=1}^{U}\delta(y_u, \hat{y}_u) \quad (14)$$

### 3.3 EXPERIMENT 1: Enhanced B-PETs

#### 3.3.1 Results

Table 2 shows the combined effect of the enhancements (EB-PETs) described in section 2.2 relative to the baseline B-PETs probability estimations. The *Win* column after each metric indicates whether the enhancements significantly improved the associated metric (*W*), hurt the metric (*L*), or did not have a significant effect (empty cells). Significance was determined by a $t$-test with level of confidence 0.1. The enhancements result in 13 wins, 5 ties and 2 losses according to the 01-MSE metric using these datasets. The average log-loss (AvLL) metric has 14/5/1 wins/ties/losses, area under the lift chart (AULC) gives 10/8/2, and the change in accuracy ($\Delta\text{Acc}$) scores 15/3/2.

Clearly the combined effect of the enhancements is very positive, but the overall comparison does not indicate which enhancements have the biggest positive impact on which metrics. To help understand the effect of each enhancement, the baseline B-PETs results are compared to results based on the individual enhancements. The summary win/tie/loss results are shown in table 3, along with the effect of incorporating all of the enhancements. Including the OB examples in the calculation of probabilities seems to have the biggest overall positive impact from the baseline. It is the only enhancement that, by itself, improves all metrics. Additionally, it results in the largest improvement in the average probability estimates, as measured by the 0/1-MSE and AvLL. Omitting smoothing also appears to have a positive effect on the average probability estimates, but has no obvious effect on AULC or $\Delta\text{Acc}$. Random feature selection, on the other hand, has a large positive effect on AULC and $\Delta\text{Acc}$, but no obvious effect on the average prediction error.

Table 3: Effect of Enhancements to B-PETs (win/tie/loss)

| Enhancement | 0/1-MSE | AvLL | AULC | $\Delta$Acc |
|---|---|---|---|---|
| OB Examples | 11/8/1 | 10/7/3 | 2/18/0 | 6/13/1 |
| Omit Smoothing | 7/11/2 | 11/5/4 | 0/20/0 | 1/18/1 |
| Random Features | 7/7/6 | 6/7/7 | 10/8/2 | 14/2/4 |
| All (EB-PETs) | 13/5/2 | 14/5/1 | 10/8/2 | 15/3/2 |

Table 4: Effect of Removing Enhancements from EB-PETs

| Enhncmt Removed | 0/1-MSE | AvLL | AULC | $\Delta$Acc |
|---|---|---|---|---|
| OB Examples | 4/10/6 | 5/9/6 | 1/19/0 | 7/6/7 |
| Omit Smoothing | 0/5/15 | 0/4/16 | 0/19/1 | 0/13/7 |
| Random Features | 6/4/10 | 7/6/7 | 2/8/10 | 2/3/15 |

The results from table 3 do not address the interaction between enhancements. To analyze those issues, a comparison is made between the EB-PETs results and results obtained by removing each enhancement individually. Table 4 summarizes the win/tie/loss results when removing each enhancement from the full EB-PETs. In this case, a higher number of losses implies the enhancement has a positive effect on the final algorithm. These results show substantially different effects of the enhancements when combined than when taken individually. However, the enhancements largely result in more losses than wins across the metrics, with the main exception being a single win on the AULC metric for the OB Examples enhancement. In the final algorithm, the omission of smoothing clearly has the most positive impact on the average prediction error of any enhancement. Random feature selection, which had no real impact on the average prediction error by itself, shows a small positive impact on the combined algorithm and still has a large positive impact on AULC and $\Delta\text{Acc}$.

#### 3.3.2 Discussion

All of the hypotheses behind the algorithm enhancements seem to be supported. The 58% increase in examples used for estimating probabilities by including OB points has a significant positive effect from the baseline, though less so in the final combined method. It appears that Bagging does provide enough diversity by itself to justify the elimination of smoothing the leaf estimates. Furthermore, the removal of smoothing seems to play an extremely important role in the performance of the final predictor. This is not surprising given the radical change in estimates that can result from typical Laplace smoothing. However, the results here do not imply that a smoothing technique can not outperform these unsmoothed estimates. In fact, it is likely that such methods can be found (c.f., Ferri et al., 2003). Alternatively, Ferri et al. also imply that pruning might be beneficial in this case, since smoothing is disabled. Finally, the diversity injected by random feature selection also appears to contribute significantly to the quality of the final predictor.

Table 2: Effect of B-PETs Enhancements (EB-PETs) by Dataset

| DS Id | 0/1-MSE B-PETs | 0/1-MSE EB-PETs | Win | Ave Log-Loss B-PETs | Ave Log-Loss EB-PETs | Win | Area under Lift Chart B-PETs | Area under Lift Chart EB-PETs | Win | Change in Accuracy B-PETs | Change in Accuracy EB-PETs | Win |
|---|---|---|---|---|---|---|---|---|---|---|---|---|
| 1 | 0.045 | 0.050 | | 0.258 | 0.342 | L | 1.683 | 1.693 | W | -0.038 | 0.005 | W |
| 2 | 0.197 | 0.197 | | 0.839 | 0.840 | | 1.376 | 1.377 | | -0.006 | 0.000 | W |
| 3 | 0.109 | 0.098 | W | 0.522 | 0.485 | W | 1.599 | 1.612 | W | -0.015 | -0.001 | W |
| 4 | 0.010 | 0.012 | L | 0.086 | 0.086 | | 1.348 | 1.348 | | -0.001 | 0.000 | |
| 5 | 0.247 | 0.069 | W | 1.822 | 0.366 | W | 1.246 | 1.632 | W | -0.171 | 0.008 | W |
| 6 | 0.0009 | 0.0005 | W | 0.043 | 0.017 | W | 1.698 | 1.698 | | 0.000 | 0.000 | |
| 7 | 0.042 | 0.038 | W | 0.246 | 0.223 | W | 1.685 | 1.688 | W | -0.007 | -0.002 | W |
| 8 | 0.212 | 0.136 | W | 0.875 | 0.631 | W | 1.553 | 1.602 | W | -0.148 | 0.049 | W |
| 9 | 0.162 | 0.160 | | 0.706 | 0.696 | W | 1.439 | 1.443 | | -0.001 | 0.001 | |
| 10 | 0.053 | 0.039 | W | 0.256 | 0.226 | W | 1.679 | 1.688 | W | -0.036 | -0.001 | W |
| 11 | 0.146 | 0.145 | | 0.649 | 0.657 | | 1.568 | 1.597 | W | -0.027 | -0.003 | W |
| 12 | 0.169 | 0.127 | W | 0.707 | 0.561 | W | 1.489 | 1.569 | W | -0.114 | 0.003 | W |
| 13 | 0.231 | 0.232 | | 0.937 | 0.939 | | 1.271 | 1.260 | L | 0.011 | -0.004 | L |
| 14 | 0.058 | 0.045 | W | 0.316 | 0.266 | W | 1.673 | 1.684 | W | 0.026 | 0.043 | W |
| 15 | 0.047 | 0.039 | W | 0.286 | 0.272 | W | 1.694 | 1.698 | | -0.023 | 0.001 | W |
| 16 | 0.033 | 0.027 | W | 0.204 | 0.154 | W | 1.633 | 1.639 | W | -0.004 | 0.000 | W |
| 17 | 0.036 | 0.034 | W | 0.191 | 0.184 | W | 1.650 | 1.651 | | -0.007 | -0.002 | W |
| 18 | 0.035 | 0.022 | W | 0.194 | 0.154 | W | 1.686 | 1.691 | W | -0.033 | 0.000 | W |
| 19 | 0.166 | 0.170 | L | 0.746 | 0.757 | | 1.242 | 1.209 | L | 0.008 | -0.010 | L |
| 20 | 0.036 | 0.017 | W | 0.267 | 0.170 | W | 1.639 | 1.639 | | -0.005 | 0.007 | W |

Table 5: Performance of MOB-ESP (wins/ties/losses)

| Competitor | 0/1-MSE | AvLL | AULC | ΔAcc |
|---|---|---|---|---|
| B-PETs | 17/3/0 | 17/3/0 | 11/6/3 | 14/4/2 |
| EB-PETs | 16/3/1 | 16/3/1 | 1/12/7 | 7/6/7 |

### 3.4 EXPERIMENT 2: MOB-ESP

#### 3.4.1 Results

In the second experiment, MOB-ESP is compared to the baseline B-PETs algorithm and to EB-PETs. Table 5 summarizes the win/tie/loss results of these comparisons and table 6 presents the full detail relative to EB-PETs for each metric and dataset. As can be seen, MOB-ESP significantly outperforms the baseline B-PETs algorithm on all metrics for these 20 datasets. It also performs much better than EB-PETs in terms of the average prediction accuracy as measured by 0/1-MSE and AvLL. EB-PETs, on the other hand, does better on AULC.

#### 3.4.2 Discussion

Again, the hypotheses behind MOB-ESP appear largely to be validated. Basing estimates on the OB classification technique and conditioning on the ensemble classification provides much larger increases in the average accuracy measures than achieved by EB-PETs. However, it is surprising that EB-PETs does so much better on the AULC metric and this is a topic for future investigation.

As noted earlier, the OB classification is based on $0.37T$ trees. Given $T$=128 in these experiments, it is unlikely for most datasets that $0.37T$ is enough trees to reach peak classification performance. Also, since it is not equal to the number of trees involved in the ensemble classification of unlabeled examples, there are almost certainly differences in classification performance between the OB examples and the unlabeled examples. This implies that better results might be achievable by constructing more trees and using only 37% of them to classify unlabeled examples.

In these experiments, $\alpha_l$ of equation 8 was set to 1. It seems probable that the OB examples are better predictors of class probabilities than the IB examples, since they were not used in the construction of the trees. This would be especially true when there are a sufficient number of OB examples. Automatically choosing a more appropriate value for $\alpha_l$ to accommodate this hypothesis is an area for future research.

The MOB-ESP estimation method can be applied in concert with a broader category of classifiers. The key prerequisite is that the classifier provide an association between an unlabeled example $\mathbf{x}_u$ and a subset of training examples that have primary responsibility for the classification of $\mathbf{x}_u$.

## 4 CONCLUSION

Combining Bagging with random feature selection has been shown to improve ensemble tree classification accuracy (Breiman, 2001), but those models do not output example-specific class probability estimates, $\hat{p}(k|\mathbf{x})$. B-PETs, on the other hand, explicitly compute class probability estimates with good rankings (Provost and Domingos, 2003). However, Margineantu and Dietterich (2002) and others have demonstrated that B-PETs do not provide particularly accurate estimates.

EB-PETs were introduced here to improve on those estimates. Experiments on 20 benchmark datasets show EB-PETs provide much more accurate average predictions according to the 0/1-MSE and AvLL metrics than do B-PETs. The experiments also suggest that EB-PETs provide bet-ter probability rankings and should be strongly considered wherever AULC is the best metric associated with a problem's solution.

Table 6: Comparison of MOB-ESP to EB-PETs by Dataset

| DS Id | 0/1-MSE | | | Ave Log-Loss | | | Area under Lift Chart | | | Change in Accuracy | | |
|---|---|---|---|---|---|---|---|---|---|---|---|---|
| | EB-PETs | MobEsp | Win | EB-PETs | MobEsp | Win | EB-PETs | MobEsp | Win | EB-PETs | MobEsp | Win |
| 1 | 0.050 | 0.004 | W | 0.342 | 0.032 | W | 1.693 | 1.691 | | 0.005 | 0.000 | L |
| 2 | 0.197 | 0.199 | | 0.840 | 0.847 | | 1.377 | 1.366 | L | 0.000 | -0.004 | L |
| 3 | 0.098 | 0.095 | W | 0.485 | 0.462 | W | 1.612 | 1.615 | | -0.001 | -0.001 | |
| 4 | 0.012 | 0.002 | W | 0.086 | 0.024 | W | 1.348 | 1.346 | | 0.000 | 0.000 | |
| 5 | 0.069 | 0.056 | W | 0.366 | 0.282 | W | 1.632 | 1.633 | | 0.008 | 0.000 | L |
| 6 | 0.0005 | 0.0000 | W | 0.017 | 0.007 | W | 1.698 | 1.698 | | 0.000 | 0.000 | |
| 7 | 0.038 | 0.026 | W | 0.223 | 0.143 | W | 1.688 | 1.688 | | -0.002 | 0.000 | W |
| 8 | 0.136 | 0.127 | W | 0.631 | 0.585 | W | 1.602 | 1.598 | | 0.049 | 0.011 | L |
| 9 | 0.16 | 0.162 | L | 0.696 | 0.705 | L | 1.443 | 1.438 | L | 0.001 | -0.002 | L |
| 10 | 0.039 | 0.031 | W | 0.226 | 0.164 | W | 1.688 | 1.686 | L | -0.001 | 0.000 | W |
| 11 | 0.145 | 0.131 | W | 0.657 | 0.595 | W | 1.597 | 1.602 | | -0.003 | 0.000 | W |
| 12 | 0.127 | 0.129 | | 0.561 | 0.576 | | 1.569 | 1.561 | | 0.003 | -0.003 | L |
| 13 | 0.232 | 0.229 | W | 0.939 | 0.928 | W | 1.260 | 1.280 | W | -0.004 | 0.004 | W |
| 14 | 0.045 | 0.039 | W | 0.266 | 0.218 | W | 1.684 | 1.680 | L | 0.043 | 0.039 | |
| 15 | 0.039 | 0.011 | W | 0.272 | 0.084 | W | 1.698 | 1.691 | L | 0.001 | 0.000 | |
| 16 | 0.027 | 0.025 | W | 0.154 | 0.142 | W | 1.639 | 1.638 | | 0.000 | 0.000 | |
| 17 | 0.034 | 0.032 | W | 0.184 | 0.169 | W | 1.651 | 1.651 | | -0.002 | 0.000 | W |
| 18 | 0.022 | 0.014 | W | 0.154 | 0.112 | W | 1.691 | 1.677 | L | 0.000 | 0.000 | W |
| 19 | 0.170 | 0.169 | | 0.757 | 0.761 | | 1.209 | 1.203 | | -0.010 | -0.004 | W |
| 20 | 0.017 | 0.008 | W | 0.170 | 0.071 | W | 1.639 | 1.629 | L | 0.007 | 0.000 | L |

This paper also introduced MOB-ESP, a technique that calculates probability estimates based on examples not used during the construction of the trees. Experiments show that this method performs significantly better than either the baseline B-PETs algorithm or the enhanced EB-PETs algorithm in regard to estimating probabilities. The experiments also suggest that the probability rankings of MOB-ESP are far better than those of B-PETs. Further work is required to determine whether MOB-ESP can be modified to perform as well as EB-PETs on the area under the lift chart metric.

If the objective is to calculate accurate probability estimates, these results strongly suggest that MOB-ESP be analyzed in the relevant domain. This is often the case in medical diagnosis, weather forecasting, economic predictions, and cost-sensitive decision making in general.